Search Methodologies: Introductory Tutorials in Optimization and Decision Support Techniques

Edmund K. Burke (Editor), Graham Kendall (Editor)

Chapter 13

# ARTIFICIAL IMMUNE SYSTEMS

U. Aickelin[#] and D. Dasgupta*

[#]*University of Nottingham, Nottingham NG8 1BB, UK*

*\*University of Memphis, Memphis, TN 38152, USA*

## 1. INTRODUCTION

The biological immune system is a robust, complex, adaptive system that defends the body from foreign pathogens. It is able to categorize all cells (or molecules) within the body as self-cells or non-self cells. It does this with the help of a distributed task force that has the intelligence to take action from a local and also a global perspective using its network of chemical messengers for communication. There are two major branches of the immune system. The innate immune system is an unchanging mechanism that detects and destroys certain invading organisms, whilst the adaptive immune system responds to previously unknown foreign cells and builds a response to them that can remain in the body over a long period of time. This remarkable information processing biological system has caught the attention of computer science in recent years.

A novel computational intelligence technique, inspired by immunology, has emerged, called Artificial Immune Systems. Several concepts from the immune have been extracted and applied for solution to real world science and engineering problems. In this tutorial, we briefly describe the immune system metaphors that are relevant to existing Artificial Immune Systems methods. We will then show illustrative real-world problems suitable for Artificial Immune Systems and give a step-by-step algorithm walkthrough for one such problem. A comparison of the Artificial Immune Systems to other well-known algorithms, areas for future work, tips & tricks and a list of resources will round this tutorial off. It should be noted that as Artificial Immune Systems is still a young and evolving field, there is not yet a fixed algorithm template and hence actual implementations might differ somewhat from time to time and from those examples given here.

## 2. OVERVIEW OF THE BIOLOGICAL IMMUNE SYSTEM

The biological immune system is an elaborate defense system which has evolved over millions of years. While many details of the immune mechanisms (innate and adaptive) and processes (humeral and cellular) are yet unknown (even to immunologists), it is, however, well-known that the immune system uses multilevel (and overlapping) defense both in parallel and sequential fashion. Depending on the type of the pathogen, and the way it gets into the body, the immune system uses different response mechanisms (differential pathways) either to neutralize the pathogenic effect or to destroy the infected cells. A detailed overview of the immune system can be found in many textbooks, for instance Kubi (2002). The immune features that are particularly relevant to our tutorial are matching, diversity and distributed control. Matching refers to the binding between antibodies and antigens. Diversity refers to the fact that, in order to achieve optimal antigen space coverage, antibody diversity must be encouraged according to Hightower et al (1995). Distributed control means that there is no central controller;

rather, the immune system is governed by local interactions among immune cells and antigens.

Two of the most important-cells in this process are white blood cells, called T-cells, and B-cells. Both of these originate in the bone marrow, but T-cells pass on to the thymus to mature, before they circulate the body in the blood and lymphatic vessels.

The T-cells are of three types; T helper cells which are essential to the activation of B-cells, Killer T-cells which bind to foreign invaders and inject poisonous chemicals into them causing their destruction, and suppressor T-cells which inhibit the action of other immune cells thus preventing allergic reactions and autoimmune diseases.

B-cells are responsible for the production and secretion of antibodies, which are specific proteins that bind to the antigen. Each B-cell can only produce one particular antibody. The antigen is found on the surface of the invading organism and the binding of an antibody to the antigen is a signal to destroy the invading cell as shown in Figure 1.

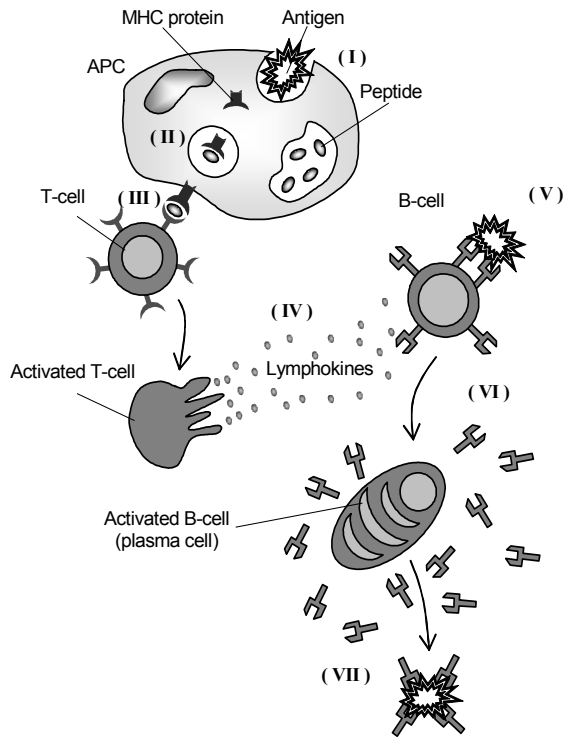

*Figure -1.* Pictorial representation of the essence of the acquired immune system mechanism (taken from de Castro and van Zuben (1999): I-II show the invade entering the body and activating T-Cells, which then in IV activate the B-cells, V is the antigen matching, VI the antibody production and VII the antigen's destruction.

As mentioned above, the human body is protected against foreign invaders by a multi-layered system. The immune system is composed of physical barriers such as the skin and respiratory system; physiological barriers such as destructive enzymes and stomach acids; and the immune system, which has can be broadly divided under two heads – Innate (non-specific) Immunity and Adaptive (specific) Immunity, which are inter-linked and influence each other. The Adaptive Immunity again is subdivided under two heads – Humoral Immunity and Cell Mediated Immunity.

Innate Immunity: The Innate Immunity is present at birth. Physiological conditions such as pH, temperature and chemical mediators provide inappropriate living conditions for foreign organisms. Also microorganisms are coated with antibodies and/or complement products (opsonization) so that they are easily recognized. Extracellular material is then ingested by macrophages by a process called phagocytosis. Also $T_{DH}$ Cells influences the phagocytosis of macrophages by secreting certain chemical messengers called lymphokines. The low levels of sialic acid on foreign antigenic surfaces make $C_3b$ bind to these surfaces for a long time and thus activate alternative pathways. Thus MAC is formed, which puncture the cell surfaces and kill the foreign invader.

Adaptive Immunity: Adaptive Immunity is the main focus of interest here as learning, adaptability, and memory are important characteristics of Adaptive Immunity. It is subdivided under two heads – Humoral Immunity and Cell Mediated Immunity.

Humoral immunity: Humoral immunity is mediated by antibodies contained in body fluids (known as humors). The humoral branch of the immune system involves interaction of B cells with antigen and their subsequent proliferation and differentiation into antibody-secreting plasma cells. Antibody functions as the effectors of the humoral response by binding to antigen and facilitating its elimination. When an antigen is coated with antibody, it can be eliminated in several ways. For example, antibody can cross-link the antigen, forming clusters that are more readily ingested by phagocytic cells. Binding of antibody to antigen on a microorganism also can activate the complement system, resulting in lysis of the foreign organism.

Cellular immunity: Cellular immunity is cell-mediated; effector T cells generated in response to antigen are responsible for cell-mediated immunity. Cytotoxic T lymphocytes (CTLs) participate in cell-mediated immune reactions by killing altered self-cells; they play an important role in the killing of virus-infected cells and tumor cells. Cytokines secreted by $T_{DH}$ can mediate the cellular immunity, and activate various phagocytic cells, enabling them to phagocytose and kill microorganisms more effectively. This type of cell-mediated immune response is especially important in host defense against intracellular bacteria and protozoa.

Whilst there is more than one mechanism at work (see Farmer (1986), Kubi (2002) or Jerne (1973) for more details), the essential process is the matching of antigen and antibody, which leads to increased concentrations (proliferation) of more closely matched antibodies. In particular, idiotypic network theory, negative selection mechanism, and the 'clonal selection' and 'somatic hypermutation' theories are primarily used in Artificial Immune Systems models.

## 2.1 Immune Network Theory

The immune Network theory had been proposed in the mid-seventies (Jerne 1974). The hypothesis was that the immune system maintains an idiotypic network of interconnected B cells for antigen recognition. These cells both stimulate and suppress each other in certain ways that lead to the stabilization of the network. Two B cells are connected if the affinities they share exceed a certain threshold, and the strength of the connection is directly proportional to the affinity they share.

## 2.2 Negative Selection mechanism

The purpose of negative selection is to provide tolerance for self cells. It deals with the immune system's ability to detect unknown antigens while not reacting to the self cells. During the generation of T-cells, receptors are made through a pseudo-random genetic rearrangement process. Then, they undergo a censoring process in the thymus, called the negative selection. There, T-cells that react against self-proteins are destroyed; thus, only those that do not bind to self-proteins are allowed to leave the thymus. These matured T-cells then circulate throughout the body to perform immunological functions and protect the body against foreign antigens.

## 2.3 Clonal Selection Principle

The clonal selection principle describes the basic features of an immune response to an antigenic stimulus. It establishes the idea that only those cells that recognize the antigen proliferate, thus being selected against those that do not. The main features of the clonal selection theory are that:

- The new cells are copies of their parents (clone) subjected to a mutation mechanism with high rates (somatic hypermutation);
- Elimination of newly differentiated lymphocytes carrying self-reactive receptors;
- Proliferation and differentiation on contact of mature cells with antigens.

When an antibody strongly matches an antigen the corresponding B-cell is stimulated to produce clones of itself that then produce more antibodies. This (hyper) mutation, is quite rapid, often as much as "one mutation per cell division" (de Castro and Von Zuben, 1999). This allows a very quick response to the antigens. It should be noted here that in the Artificial

Immune Systems literature, often no distinction is made between B-cells and the antibodies they produce. Both are subsumed under the word 'antibody' and statements such as mutation of antibodies (rather than mutation of B-cells) are common.

There are many more features of the immune system, including adaptation, immunological memory and protection against auto-immune attacks, not discussed here. In the following sections, we will revisit some important aspects of these concepts and show how they can be modelled in 'artificial' immune systems and then used to solve real-world problems. First, let us give an overview of typical problems that we believe are amenable to being solved by Artificial Immune Systems.

## 3. ILLUSTRATIVE PROBLEMS

### 3.1 Intrusion Detection Systems

Anyone keeping up-to-date with current affairs in computing can confirm numerous cases of attacks made on computer servers of well-known companies. These attacks range from denial-of-service attacks to extracting credit-card details and sometimes we find ourselves thinking "haven't they installed a firewall"? The fact is they often have a firewall. A firewall is a useful, often essential, but current firewall technology is insufficient to detect and block all kinds of attacks.

However, on ports that need to be open to the internet, a firewall can do little to prevent attacks. Moreover, even if a port is blocked from internet access, this does not stop an attack from inside the organisation. This is where Intrusion Detection Systems come in. As the name suggests, Intrusion Detection Systems are installed to identify (potential) attacks and to react by usually generating an alert or blocking the unscrupulous data.

The main goal of Intrusion Detection Systems is to detect unauthorised use, misuse and abuse of computer systems by both system insiders and external intruders. Most current Intrusion Detection Systems define suspicious signatures based on known intrusions and probes. The obvious limit of this type of Intrusion Detection Systems is its failure of detecting previously unknown intrusions. In contrast, the human immune system adaptively generates new immune cells so that it is able to detect previously unknown and rapidly evolving harmful antigens (Forrest et al 1994). Thus the challenge is to emulate the success of the natural systems.

## 3.2 Data Mining – Collaborative Filtering and Clustering

Collaborative Filtering is the term for a broad range of algorithms that use similarity measures to obtain recommendations. The best-known example is probably the "people who bought this also bought" feature of the internet company Amazon (2003). However, any problem domain where users are required to rate items is amenable to Collaborative Filtering techniques. Commercial applications are usually called recommender systems (Resnick and Varian 1997). A canonical example is movie recommendation.

In traditional Collaborative Filtering, the items to be recommended are treated as 'black boxes'. That is, your recommendations are based purely on the votes of other users, and not on the content of the item. The preferences of a user, usually a set of votes on an item, comprise a user profile, and these profiles are compared in order to build a neighbourhood. The key decision is what similarity measure is used: The most common method to compare two users is a correlation-based measure like Pearson or Spearman, which gives two neighbours a matching score between -1 and 1. The canonical example is the k-Nearest-Neighbour algorithm, which uses a matching method to select k reviewers with high similarity measures. The votes from these reviewers, suitably weighted, are used to make predictions and recommendations.

The evaluation of a Collaborative Filtering algorithm usually centres on its accuracy. There is a difference between prediction (given a movie, predict a given user's rating of that movie) and recommendation (given a user, suggest movies that are likely to attract a high rating). Prediction is easier to assess quantitatively but recommendation is a more natural fit to the movie domain. A related problem to Collaborative Filtering is that of clustering data or users in a database. This is particularly useful in very large databases, which have become too large to handle. Clustering works by dividing the entries of the database into groups, which contain people with similar preferences or in general data of similar type.

# 4. ARTIFICIAL IMMUNE SYSTEMS BASIC CONCEPTS

## 4.1 Initialisation / Encoding

To implement a basic Artificial Immune System, four decisions have to be made: Encoding, Similarity Measure, Selection and Mutation. Once an encoding has been fixed and a suitable similarity measure is chosen, the algorithm will then perform selection and mutation, both based on the similarity measure, until stopping criteria are met. In this section, we will describe each of these components in turn.

Along with other heuristics, choosing a suitable encoding is very important for the algorithm's success. Similar to Genetic Algorithms, there is close inter-play between the encoding and the fitness function (the later is in Artificial Immune Systems referred to as the 'matching' or 'affinity' function). Hence both ought to be thought about at the same time. For the current discussion, let us start with the encoding.

First, let us define what we mean by 'antigen' and 'antibody' in the context of an application domain. Typically, an antigen is the target or solution, e.g. the data item we need to check to see if it is an intrusion, or the user that we need to cluster or make a recommendation for. The antibodies are the remainder of the data, e.g. other users in the data base, a set of network traffic that has already been identified etc. Sometimes, there can be more than one antigen at a time and there are usually a large number of antibodies present simultaneously.

Antigens and antibodies are represented or encoded in the same way. For most problems the most obvious representation is a string of numbers or features, where the length is the number of variables, the position is the variable identifier and the value (could be binary or real) of the variable. For instance, in a five variable binary problem, an encoding could look like this: (10010).

As mentioned previously, for data mining and intrusion detection applications. What would an encoding look like in these cases? For data mining, let us consider the problem of recommending movies. Here the encoding has to represent a user's profile with regards to the movies he has seen and how much he has (dis)liked them. A possible encoding for this could be a list of numbers, where each number represents the 'vote' for an item. Votes could be binary (e.g. Did you visit this web page?), but can also be integers in a range (say [0, 5], i.e. 0 - did not like the movie at all, 5 – did like the movie very much).

Hence for the movie recommendation, a possible encoding is:

$$User = \{\{id_1, score_1\}, \{id_2, score_2\}...\{id_n, score_n\}\}$$

Where *id* corresponds to the unique identifier of the movie being rated and score to this user's score for that movie. This captures the essential features of the data available (Cayzer and Aickelin 2002).

For intrusion detection, the encoding may be to encapsulate the essence of each data packet transferred, e.g. [<protocol> <source ip> <source port> <destination ip> <destination port>], example: [<tcp> <113.112.255.254> <108.200.111.12> <25> which represents an incoming data packet send to port 25. In these scenarios, wildcards like 'any port' are also often used.

## 4.2    Similarity or Affinity Measure

As mentioned in the previous section, similarity measure or matching rule is one of the most important design choices in developing an Artificial Immune Systems algorithm, and is closely coupled to the encoding scheme.

Two of the simplest matching algorithms are best explained using binary encoding: Consider the strings (00000) and (00011). If one does a bit-by-bit comparison, the first three bits are identical and hence we could give this pair a matching score of 3. In other words, we compute the opposite of the Hamming Distance (which is defined as the number of bits that have to be changed in order to make the two strings identical).

Now consider this pair: (00000) and (01010). Again, simple bit matching gives us a similarity score of 3. However, the matching is quite different as the three matching bits are not connected. Depending on the problem and encoding, this might be better or worse. Thus, another simple matching algorithm is to count the number of continuous bits that match and return the length of the longest matching as the similarity measure. For the first example above this would still be 3, for the second example this would be 1.

If the encoding is non-binary, e.g. real variables, there are even more possibilities to compute the 'distance' between the two strings, for instance we could compute the geometrical (Euclidian) distance etc.

For data mining problems, like the movie recommendation system, similarity often means 'correlation'. Take the movie recommendation problem as an example and assume that we are trying to find users in a database that are similar to the key user who's profile were are trying to match in order to make recommendations. In this case, what we are trying to measure is how similar are the two users' tastes. One of the easiest ways of doing this is to compute the Pearson Correlation Coefficient between the two users.

I.e. if the Pearson measure is used to compare two user's u and v:

$$r = \frac{\sum_{i=1}^{n}(u_i - \bar{u})(v_i - \bar{v})}{\sqrt{\sum_{i=1}^{n}(u_i - \bar{u})^2 \sum_{i=1}^{n}(v_i - \bar{v})^2}} \quad (1)$$

Where u and v are users, n is the number of overlapping votes (i.e. Movies for which both u and v have voted), $u_i$ is the vote of user u for movie i and ū is the average vote of user u over all films (not just the overlapping votes). The measure is amended so default to a value of 0 if the two users have no films in common. During our research reported in Cayzer and Aickelin (2002a, 2002b) we also found it useful to introduce a penalty parameter (c.f. penalties in genetic algorithms) for users who only have very few films in common, which in essence reduces their correlation.

The outcome of this measure is a value between -1 and 1, where values close to 1 mean strong agreement, values near to -1 mean strong disagreement and values around 0 mean no correlation. From a data mining point of view, those users who score either 1 or -1 are the most useful and hence will be selected for further treatment by the algorithm.

For other applications, 'matching' might not actually be beneficial and hence those items that match might be eliminated. This approach is known as 'negative selection' and mirrors what is believed to happen during the maturation of B-cells who have to learn not to 'match' our own tissues as otherwise we would be subject to auto-immune diseases.

Under what circumstance would a negative selection algorithm be suitable for an Artificial Immune Systems implementation? Consider the case of Intrusion Detection as solved by Hofmeyr and Forrest (2000). One way of solving this problem is by defining a set of 'self', i.e. a trusted network, our company's computers, known partners etc. During the initialisation of the algorithm, we would then randomly create a large number of so called 'detectors', i.e. strings that looks similar to the sample Intrusion Detection Systems encoding given above. We would then subject these detectors to a matching algorithm that compares them to our 'self'. Any matching detector would be eliminated and hence we select those that do no match (negative selection). All non-matching detectors will then form our final detector set. This detector set is then used in the second phase of the algorithm to continuously monitor all network traffic. Should a match be found now the algorithm would report this as a possible alert or 'non-self'. There are a number of problems with this approach, which we shall discuss further in the Enhancements and Future Application Section.

### 4.3    Negative, Clonal or Neighbourhood Selection

The meaning of this step differs somewhat depending on the exact problem the Artificial Immune Systems is applied to. We have already described the concept of negative selection above. For the film recommender, choosing a suitable neighbourhood means choosing good correlation scores and hence we will perform 'positive' selection. How would the algorithm use this?

Consider the Artificial Immune Systems to be empty at the beginning. The target user is encoded as the antigen, and all other users in the database are possible antibodies. We add the antigen to the Artificial Immune Systems and then we add one candidate antibody at a time. Antibodies will start with a certain concentration value. This value is decreasing over time (death rate), similar to the evaporation in Ant Systems. Antibodies with a sufficiently low concentration are removed from the system, whereas antibodies with a high concentration may saturate. However, an antibody can increase its concentration by matching the antigen, the better the match the higher the increase (a process called 'stimulation'). The process of stimulation or increasing concentration can also be regarded as 'cloning' if one thinks in a discrete setting. Once enough antibodies have been added to the system, it starts to iterate a loop of reducing concentration and stimulation until at least one antibody drops out. A new antibody is added and the process repeated until the Artificial Immune Systems is stabilised, i.e. there are no more drop-outs for a certain period of time.

Mathematically, at each step (iteration) an antibody's concentration is increased by an amount dependent on its matching to each antigen. In absence of matching, an antibody's concentration will slowly decrease over time. Hence an Artificial Immune Systems iteration is governed by the following equation, based on Farmer et al (1986):

$$\frac{dx_i}{dt} = \left[\binom{antigens}{recognised} - \binom{death}{rate}\right]$$

$$= \left[k_2(\sum_{j=1}^{N} m_{ji} x_i y_j) - k_3 x_i\right]$$

Where:
N is the number of antigens.
$x_i$ is the concentration of antibody i
$y_j$ is the concentration of antigen j
$k_2$ is the stimulation effect and $k_3$ is the death rate
$m_{ji}$ is the matching function between antibody i & antibody (or antigen) j

The following pseudo code summarise the Artificial Immune Systems of the movie recommender:

Initialise Artificial Immune Systems
Encode user for whom to make predictions as antigen Ag
WHILE (Artificial Immune Systems not Full) & (More Antibodies) DO
    Add next user as an antibody Ab
    Calculate matching scores between Ab and Ag
    WHILE (Artificial Immune Systems at full size) & (Artificial Immune Systems not Stabilised) DO
        Reduce Concentration of all Abs by a fixed amount
        Match each Ab against Ag and stimulate as necessary
    OD
OD
Use final set of Antibodies to produce recommendation.

In this example, the Artificial Immune Systems is considered stable after iterating for ten iterations without changing in size. Stabilisation thus means that a sufficient number of 'good' neighbours have been identified and therefore a prediction can be made. 'Poor' neighbours would be expected to drop out of the Artificial Immune Systems after a few iterations. Once the Artificial Immune Systems has stabilised using the above algorithm, we use the antibody concentration to weigh the neighbours and then perform a weighted average type recommendation.

## 4.4   Somatic Hypermutation

The mutation most commonly used in Artificial Immune Systems is very similar to that found in Genetic Algorithms, e.g. for binary strings bits are flipped, for real value strings one value is changed at random, or for others the order of elements is swapped. In addition, the mechanism is often enhanced by the 'somatic' idea, i.e. the closer the match (or the less close the match, depending on what we are trying to achieve), the more (or less) disruptive the mutation.

However, mutating the data might not make sense for all problems considered. For instance, it would not be suitable for the movie recommender. Certainly, mutation could be used to make users more similar to the target, however, the validity of recommendations based on these artificial users is questionable and if over-done, we would end up with the target user itself. Hence for some problems, somatic Hypermutation is not used, since it is not immediately obvious how to mutate the data sensibly such that these artificial entities still represent plausible data.

Nevertheless, for other problem domains, mutation might be very useful. For instance, taking the negative selection approach to intrusion detection, rather than throwing away matching detectors in the first phase of the algorithm, these could be mutated to safe time and effort. Also, depending on the degree of matching the mutation could e more or less strong. This was in fact one extension implemented by Hofmeyr and Forrest (2000).

For data mining problems, mutation might also be useful, if for instance the aim is to cluster users. Then the centre of each cluster (the antibodies) could be an artificial pseudo user that can be mutated at will until the desired degree of matching between the centre and antigens in its cluster is reached. This is an approach implemented by Castro and von Zuben (2001).

# 5. COMPARISON OF ARTIFICIAL IMMUNE SYSTEMS TO GENETIC ALGORITHMS AND NEURAL NETWORKS

Going through the tutorial so far, you might already have noticed that both Genetic Algorithms and Neural Networks have been mentioned a number of times. In fact, they both have a number of ideas in common with Artificial Immune Systems and the purpose of the following, self-explanatory table, is to put their similarities and differences next to each other (see Dasgupta 1999). Evolutionary computation shares many elements, concepts like population, genotype phenotype mapping, and proliferation of the most fitted are present in different Artificial Immune Systems methods.

Artificial Immune Systems models based on immune networks resembles the structures and interactions of connectionist models. Some works have pointed out the similarities and the differences between Artificial Immune Systems and artificial neural networks (Dasgupta 1999 and De Castro and Von Zuben 2001). De Castro has also used Artificial Immune Systems to initialize the centres of radial basis function neural networks and to produce a good initial set of weights for feed-forward neural networks.

It should be noted that some of the items in table 1 are gross simplifications, both to benefit the design of the table and not to overwhelm the reader. Some of these points are debatable; however, we believe that this comparison is valuable nevertheless to show exactly where Artificial Immune Systems fit in. The comparisons are based on a Genetic Algorithm (GA) used for optimisation and a Neural Network (NN) used for Classification.

|  | GA (Optimisation) | NN (Classification) | Artificial Immune Systems |
|---|---|---|---|
| Components | Chromosome Strings | Artificial Neurons | Attribute Strings |
| Location of Components | Dynamic | Pre-Defined | Dynamic |
| Structure | Discrete Components | Networked Components | Discrete components / Networked Components |
| Knowledge Storage | Chromosome Strings | Connection Strengths | Component Concentration / Network |

| | | | Connections |
|---|---|---|---|
| Dynamics | Evolution | Learning | Evolution / Learning |
| Meta-Dynamics | Recruitment / Elimination of Components | Construction / Pruning of Connections | Recruitment / Elimination of Components |
| Interaction between Components | Crossover | Network Connections | Recognition / Network Connections |
| Interaction with Environment | Fitness Function | External Stimuli | Recognition / Objective Function |
| Threshold Activity | Crowding / Sharing | Neuron Activation | Component Affinity |

Table 1: Comparison of Artificial Immune Systems to Genetic Algorithms and Neural Networks.

# 6. EXTENSIONS OF ARTIFICIAL IMMUNE SYSTEMS

## 6.1 Idiotypic Networks - Network Interactions (Suppression)

The idiotypic effect builds on the premise that antibodies can match other antibodies as well as antigens. It was first proposed by Jerne (1973) and formalised into a model by Farmer et al (1986). The theory is currently debated by immunologists, with no clear consensus yet on its effects in the humoral immune system (Kuby 2002). The idiotypic network hypothesis builds on the recognition that antibodies can match other antibodies as well as antigens. Hence, an antibody may be matched by other antibodies, which in turn may be matched by yet other antibodies. This activation can continue to spread through the population and potentially has much explanatory power. It could, for example, help explain how the memory of past infections is maintained. Furthermore, it could result in the suppression of similar antibodies thus encouraging diversity in the antibody pool. The idiotypic network has been formalised by a number of theoretical immunologists (Perelson and Weisbuch 1997):

$$\begin{aligned}\frac{dx_i}{dt} &= c\left[\binom{antibodies}{recognised} - \binom{I\ am}{recognised} + \binom{antigens}{recognised}\right] - \binom{death}{rate} \\ &= c\left[\sum_{j=1}^{N} m_{ji}x_ix_j - k_1\sum_{j=1}^{N} m_{ij}x_ix_j + \sum_{j=1}^{n} m_{ji}x_iy_j\right] - k_2 x_i \quad (1)\end{aligned}$$

Where:
$N$ is the number of antibodies and $n$ is the number of antigens.
$x_i$ (or $x_j$) is the concentration of antibody $i$ (or $j$)
$y_j$ is the concentration of antigen $j$
$c$ is a rate constant
$k_1$ is a suppressive effect and $k_2$ is the death rate
$m_{ji}$ is the matching function between antibody $i$ & antibody (or antigen) $j$

As can be seen from the above equation, the nature of an idiotypic interaction can be either positive or negative. Moreover, if the matching function is symmetric, then the balance between "I am recognised" and "Antibodies recognised" (parameters $c$ and $k_1$ in the equation) wholly determines whether the idiotypic effect is positive or negative, and we can simplify the equation. We can simplify equation (1) above still if we only allow one antigen in the Artificial Immune Systems. In the new equation (2), the first term is simplified as we only have one antigen, and the suppression term is normalised to allow a 'like for like' comparison between the different rate constants. The simplified equation looks like this:

$$\frac{dx_i}{dt} = k_1 m_i x_i y - \frac{k_2}{n}\sum_{j=1}^{n} m_{ij}x_ix_j - k_3 x_i \quad (2)$$

Where:
$k_1$ is stimulation, $k_2$ suppression and $k_3$ death rate
$m_i$ is the correlation between antibody $i$ and the (sole) antigen
$x_i$ (or $x_j$) is the concentration of antibody $i$ (or $j$)
$y$ is the concentration of the (sole) antigen
$m_{ij}$ is the correlation between antibodies $i$ and $j$
$n$ is the number of antibodies.

Why would we want to use the idotypic effect? Because it might provide us with a way of achieving 'diversity', similar to 'crowding' or 'fitness sharing' in a genetic algorithm. For instance, in the movie recommender, we want to ensure that the final neighbourhood population is diverse, so that we get more interesting recommendations. Hence, to use the idiotypic effect in the movie recommender system mentioned previously, the pseudo code would be amended by adding the following lines in italic.

```
Initialise Artificial Immune Systems
Encode user for whom to make predictions as antigen Ag
WHILE (Artificial Immune Systems not Full) & (More Antibodies) DO
   Add next user as an antibody Ab
    Calculate matching scores between Ab and Ag and Ab and other Abs
       WHILE (Artificial Immune Systems at full size) & (Artificial Immune Systems not Stabilised) DO
          Reduce Concentration of all Abs by a fixed amount
          Match each Ab against Ag and stimulate as necessary
          Match each Ab against each other Ab and execute idiotypic effect
       OD
OD
Use final set of Antibodies to produce recommendation.
```

The diagrams in figure 3 below show the idiotypic effect using dotted arrows whereas standard stimulation is shown using black arrows. In the left diagram antibodies Ab1 and Ab3 are very similar and they would have their concentrations reduced in the 'Iterate Artificial Immune Systems' stage of the algorithm above. However, in the right diagram, the four antibodies are well separated from each other as well as being close to the antigen and so would have their concentrations increased.

At each iteration of the film recommendation Artificial Immune Systems the concentration of the antibodies is changed according to the formula given on the next page. This will increase the concentration of antibodies that are similar to the antigen and can allow either the stimulation, suppression, or both, of antibody-antibody interactions to have an effect on the antibody concentration. More detailed discussion of these effects on recommendation problems are contained within Cayzer and Aickelin (2002a and 2002b).

19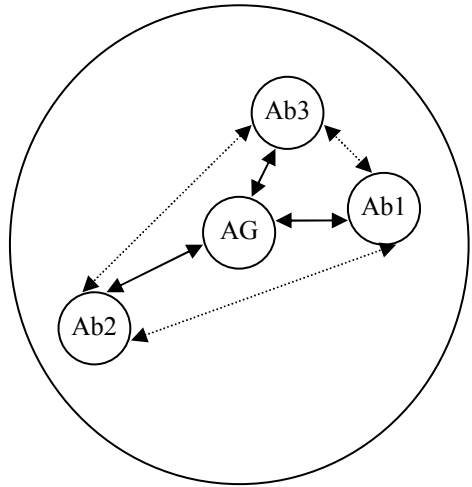

*Figure -2*. Illustration of the idiotypic effect

## 6.2 Danger Theory

Over the last decade, a new theory has become popular amongst immunologists. It is called the Danger Theory, and its chief advocate is Matzinger (1994, 2001 and 2003). A number of advantages are claimed for this theory; not least that it provides a method of 'grounding' the immune response. The theory is not complete, and there are some doubts about how much it actually changes behaviour and / or structure. Nevertheless, the theory contains enough potentially interesting ideas to make it worth assessing its relevance to Artificial Immune Systems.

However, it is not simply a question of matching in the humoral immune system. It is fundamental that only the 'correct' cells are matched as otherwise this could lead to a self-destructive autoimmune reaction. Classical immunology (Kuby 2002) stipulates that an immune response is triggered when the body encounters something non-self or foreign. It is not yet fully understood how this self-non-self discrimination is achieved, but many immunologists believe that the difference between them is learnt early in life. In particular it is thought that the maturation process plays an important role to achieve self-tolerance by eliminating those T and B-cells that react to self. In addition, a 'confirmation' signal is required; that is, for either B-cell or T (killer) cell activation, a T (helper) lymphocyte must also be activated. This dual activation is further protection against the chance of accidentally reacting to self.

Matzinger's Danger Theory debates this point of view (for a good introduction, see Matzinger 2003). Technical overviews can be found in Matzinger (1994) and Matzinger (2001). She points out that there must be discrimination happening that goes beyond the self-non-self distinction described above. For instance:

- There is no immune reaction to foreign bacteria in the gut or to the food we eat although both are foreign entities.
- Conversely, some auto-reactive processes are useful, for example against self molecules expressed by stressed cells.
- The definition of self is problematic – realistically, self is confined to the subset actually seen by the lymphocytes during maturation.
- The human body changes over its lifetime and thus self changes as well. Therefore, the question arises whether defences against non-self learned early in life might be autoreactive later.

Other aspects that seem to be at odds with the traditional viewpoint are autoimmune diseases and certain types of tumours that are fought by the immune system (both attacks against self) and successful transplants (no attack against non-self).

Matzinger concludes that the immune system actually discriminates "some self from some non-self". She asserts that the Danger Theory introduces not just new labels, but a way of escaping the semantic difficulties with self and non-self, and thus provides grounding for the immune response. If we accept the Danger Theory as valid we can take care of 'non-self but harmless' and of 'self but harmful' invaders into our system. To see how this is possible, we will have to examine the theory in more detail.

The central idea in the Danger Theory is that the immune system does not respond to non-self but to danger. Thus, just like the self-non-self theories, it fundamentally supports the need for discrimination. However, it differs in the answer to what should be responded to. Instead of responding to foreignness, the immune system reacts to danger. This theory is borne out of the observation that there is no need to attack everything that is foreign, something that seems to be supported by the counter examples above. In this theory, danger is measured by damage to cells indicated by distress signals that are sent out when cells die an unnatural death (cell stress or lytic cell death, as opposed to programmed cell death, or apoptosis).

Figure 4 depicts how we might picture an immune response according to the Danger Theory (Aickelin and Cayzer (2002)). A cell that is in distress sends out an alarm signal, where upon antigens in the neighbourhood are captured by antigen-presenting cells such as macrophages, which then travel to the local lymph node and present the antigens to lymphocytes. Essentially, the danger signal establishes a danger zone around itself. Thus B-cells producing antibodies that match antigens within the danger zone get stimulated and undergo the clonal expansion process. Those that do not match or are too far away do not get stimulated.

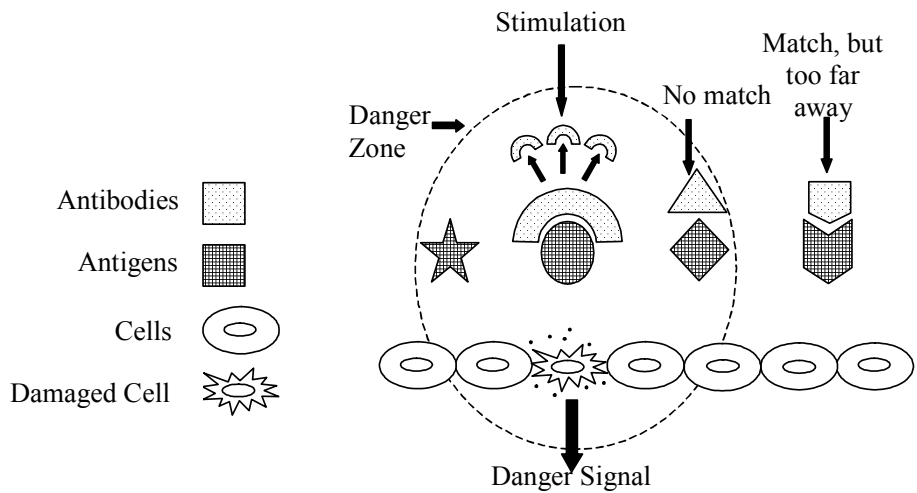

*Figure -4.* Danger Theory Illustration

Matzinger admits that the exact nature of the danger signal is unclear. It may be a 'positive' signal (for example heat shock protein release) or a 'negative' signal (for example lack of synaptic contact with a dendritic antigen-presenting cell). This is where the Danger Theory shares some of the problems associated with traditional self-non-self discrimination (i.e. how to discriminate danger from non-danger). However, in this case, the signal is grounded rather than being some abstract representation of danger.

How could we use the Danger Theory in Artificial Immune Systems? The Danger Theory is not about the way Artificial Immune Systems represent data (Aickelin and Cayzer 2002). Instead, it provides ideas about which data the Artificial Immune Systems should represent and deal with.

They should focus on dangerous, i.e. interesting data. It could be argued that the shift from non-self to danger is merely a symbolic label change that achieves nothing. We do not believe this to be the case, since danger is a grounded signal, and non-self is (typically) a set of feature vectors with no further information about whether all or some of these features are required over time. The danger signal helps us to identify which subset of feature vectors is of interest. A suitably defined danger signal thus overcomes many of the limitations of self-non-self selection. It restricts the domain of non-self to a manageable size, removes the need to screen against all self, and deals adaptively with scenarios where self (or non-self) changes over time.

The challenge is clearly to define a suitable danger signal, a choice that might prove as critical as the choice of fitness function for an evolutionary algorithm. In addition, the physical distance in the biological system should be translated into a suitable proxy measure for similarity or causality in Artificial Immune Systems. This process is not likely to be trivial. Nevertheless, if these challenges are met, then future Artificial Immune Systems applications might derive considerable benefit, and new insights, from the Danger Theory, in particular Intrusion Detection Systems.

## 7. SOME PROMISING AREAS FOR FUTURE APPLICATION

It seems intuitively obvious, that Artificial Immune Systems should be most suitable for computer security problems. If the human immune system keeps our body alive and well, why can we not do the same for computers using Artificial Immune Systems?

Earlier, we have outlined the traditional approach to do this: However, in order to provide viable Intrusion Detection Systems, Artificial Immune Systems must build a set of detectors that accurately match antigens. In current Artificial Immune Systems based Intrusion Detection Systems (Dasgupta and Gonzalez (2002), Esponda et al (2002), Hofmeyr and Forrest (2000)), both network connections and detectors are modelled as strings. Detectors are randomly created and then undergo a maturation phase where they are presented with good, i.e. self, connections. If the detectors match any of these they are eliminated otherwise they become mature. These mature detectors start to monitor new connections during their lifetime. If these mature detectors match anything else, exceeding a certain threshold value, they become activated. This is then reported to a human operator who decides whether there is a true anomaly. If so, the detectors are promoted to memory detectors with an indefinite life span and minimum activation threshold (immunisation) (Kim and Bentley 2002).

An approach such as the above is known as negative selection as only those detectors (antibodies) that do not match live on (Forrest et al. 1994). Earlier versions of negative selection algorithm used binary representation scheme, however, this scheme shows scaling problems when it is applied to real network traffic (Kim and Bentley 2001). As the systems to be protected grow larger and larger so does self and nonself. Hence, it becomes more and more problematic to find a set of detectors that provides adequate coverage, whilst being computationally efficient. It is inefficient, to map the entire self or nonself universe, particularly as they will be changing over time and only a minority of nonself is harmful, whilst some self might cause damage (e.g. internal attack). This situation is further aggravated by the fact that the labels self and nonself are often ambiguous and even with expert knowledge they are not always applied correctly (Kim and Bentley 2002).

How could this problem be overcome? One way could be to borrowed ideas from the Danger Theory to provide a way of grounding the response and hence removing the necessity to map self or nonself. In our system, the correlation of low-level alerts (danger signals) will trigger a reaction. An important and recent research issue for Intrusion Detection Systems is how to find true intrusion alerts from thousands and thousands of false alerts generated (Hofmeyr and Forrest 2000). Existing Intrusion Detection Systems employ various types of sensors that monitor low-level system events. Those sensors report anomalies of network traffic patterns, unusual terminations of UNIX processes, memory usages, the attempts to access unauthorised files, etc. (Kim and Bentley 2001). Although these reports are useful signals of real intrusions, they are often mixed with false alerts and their unmanageable volume forces a security officer to ignore most alerts (Hoagland and Staniford 2002). Moreover, the low level of alerts makes it very hard for a security officer to identify advancing intrusions that usually consist of different stages of attack sequences. For instance, it is well known that computer hackers use a number of preparatory stages (rArtificial Immune Systemsing low-level alerts) before actual hacking according to Hoaglandand and Staniford. Hence, the correlations between intrusion alerts from different attack stages provide more convincing attack scenarios than detecting an intrusion scenario based on low-level alerts from individual stages. Furthermore, such scenarios allow the Intrusion Detection Systems to detect intrusions early before damage becomes serious.

To correlate Intrusion Detection Systems alerts for detection of an intrusion scenario, recent studies have employed two different approaches: a probabilistic approach (Valdes and Skinner (2001)) and an expert system approach (Ning et al (2002)). The probabilistic approach represents known intrusion scenarios as Bayesian networks. The nodes of Bayesian networks are Intrusion Detection Systems alerts and the posterior likelihood between nodes is updated as new alerts are collected. The updated likelihood can lead to conclusions about a specific intrusion scenario occurring or not. The expert system approach initially builds possible intrusion scenarios by identifying low-level alerts. These alerts consist of prerequisites and consequences, and they are represented as hypergraphs. Known intrusion scenarios are detected by observing the low-level alerts at each stage, but these approaches have the following problems according to Cuppens et al (2002):

- Handling unobserved low-level alerts that comprise an intrusion scenario.
- Handling optional prerequisite actions.
- Handling intrusion scenario variations.

The common trait of these problems is that the Intrusion Detection Systems can fail to detect an intrusion when an incomplete set of alerts comprising an intrusion scenario is reported. In handling this problem, the probabilistic approach is somewhat more advantageous than the expert system approach because in theory it allows the Intrusion Detection Systems to correlate missing or mutated alerts. The current probabilistic approach builds Bayesian networks based on the similarities between selected alert features. However, these similarities alone can fail to identify a causal relationship between prerequisite actions and actual attacks if pairs of prerequisite actions and actual attacks do not appear frequently enough to be reported. Attackers often do not repeat the same actions in order to disguise their attempts. Thus, the current probabilistic approach fails to detect intrusions that do not show strong similarities between alert features but have causal relationships leading to final attacks. This limit means that such Intrusion Detection Systems fail to detect sophisticated intrusion scenarios.

We propose Artificial Immune Systems based on Danger Theory ideas that can handle the above Intrusion Detection Systems alert correlation problems. The Danger Theory explains the immune response of the human body by the interaction between Antigen Presenting Cells and various signals. The immune response of each Antigen Presenting Cell is determined by the generation of danger signals through cellular stress or cell death. In particular, the balance and correlation between different danger signals depending on different-cell death causes would appear to be critical to the immunological outcome. The investigation of this hypothesis is the main research goal of the immunologists for this project. The wet experiments of this project focus on understanding how the Antigen Presenting Cells react to the balance of different types of signals, and how this reaction leads to an overall immune response. Similarly, our Intrusion Detection Systems investigation will centre on understanding how intrusion scenarios would be detected by reacting to the balance of various types of alerts. In the Human Immune System, Antigen Presenting Cells activate according to the balance of apoptotic and necrotic cells and this activation leads to protective immune responses. Similarly, the sensors in Intrusion Detection Systems report various low-level alerts and the correlation of these alerts will lead to the construction of an intrusion scenario.

## 8. TRICKS OF THE TRADE

Are Artificial Immune Systems suitable for pure optimisation?

Depending on what is meant by optimisation, the answer is probably no, in the same sense as 'pure' genetic algorithms are not 'function optimizers'. One has to keep in mind that although the immune system is about matching and survival, it is really a team effort where multiple solutions are produced all the time that together provide the answer. Hence, in our opinion Artificial Immune Systems is probably more suited as an optimiser where multiple solutions are of benefit, either directly, e.g. because the problem has multiple objectives or indirectly, e.g. when a neighbourhood of solution sis produced that is then used to generate the desired outcome. However, Artificial Immune Systems can be made into more focused optimisers by adding hill-climbing or other functions that exploit local or problem specific knowledge, similar to the idea of augmenting genetic algorithm to memetic algorithms.

What problems are Artificial Immune Systems most suitable for?

As mentioned in the previous paragraph, we believe that although using Artificial Immune Systems for pure optimisation, e.g. the Travelling Salesman Problem or Job Shop Scheduling, can be made to work, this is probably missing the point. Artificial Immune Systems are powerful when a population of solution is essential either during the search or as an outcome. Furthermore, the problem has to have some concept of 'matching'. Finally, because at their heart Artificial Immune Systems are evolutionary algorithms, they are more suitable for problems that change over time rather and need to be solved again and again, rather than one-off optimisations.

Hence, the evidence seems to point to Data Mining in its wider meaning as the best area for Artificial Immune Systems.

How do I set the parameters?

Unfortunately, there is no short answer to this question. As with the majority of other heuristics that require parameters to operate, their setting is individual to the problem solved and universal values are not available. However, it is fair to say that along with other evolutionary algorithms Artificial Immune Systems are robust with respect to parameter values as long as they are chosen from a sensible range.

Why not use a Genetic Algorithm instead?

Because you may miss out on the benefits of the idiotypic network effects.

Why not use a Neural Network instead?

Because you may miss out on the benefits of a population of solutions and the evolutionary selection pressure and mutation.

Are Artificial Immune Systems Learning Classifier Systems under a different name?

No, not quite. However, to our knowledge Learning Classifier Systems are probably the most similar of the better known meta-heuristic, as they also combine some features of Evolutionary Algorithms and Neural Networks. However, these features are different. Someone who is interested in implementing and Artificial Immune Systems or Learning Classifier Systems is likely to be well advised to read about both approaches to see which one is most suited for the problem at hand.

## 9. CONCLUSIONS

The immune system is highly distributed, highly adaptive, self-organising in nature, maintains a memory of past encounters, and has the ability to continually learn about new encounters. The Artificial Immune Systems is an example of a system developed around the current understanding of the immune system. It illustrates how an Artificial Immune Systems can capture the basic elements of the immune system and exhibit some of its chief characteristics.

Artificial Immune Systems can incorporate many properties of natural immune systems, including diversity, distributed computation, error tolerance, dynamic learning and adaptation and self-monitoring. The human immune system has motivated scientists and engineers for finding powerful information processing algorithms that has solved complex engineering tasks. The Artificial Immune Systems is a general framework for a distributed adaptive system and could, in principle, be applied to many domains. Artificial Immune Systems can be applied to classification problems, optimisation tasks and other domains. Like many biologically inspired systems it is adaptive, distributed and autonomous. The primary advantages of the Artificial Immune Systems are that it only requires positive examples, and the patterns it has learnt can be explicitly examined. In addition, because it is self-organizing, it does not require effort to optimize any system parameters.

To us, the attraction of the immune system is that if an adaptive pool of antibodies can produce 'intelligent' behaviour, can we harness the power of this computation to tackle the problem of preference matching, recommendation and intrusion detection? Our conjecture is that if the concentrations of those antibodies that provide a better match are allowed to increase over time, we should end up with a subset of good matches. However, we are not interested in optimising, i.e. in finding the one best match. Instead, we require a set of antibodies that are a close match but which are at the same time distinct from each other for successful recommendation. This is where we propose to harness the idiotypic effects of binding antibodies to similar antibodies to encourage diversity.

## 10.  SOURCES OF ADDITIONAL INFORMATION

The following websites, books and proceedings should be an excellent starting point for those readers wishing to learn more about Artificial Immune Systems.

- Artificial Immune Systems and Their Applications by D Dasgupta (Editor), Springer Verlag, 1999.
- Artificial Immune Systems: A New Computational Intelligence Approach by L de Castro, J Timmis, Springer Verlag, 2002.
- Immunocomputing: Principles and Applications by A Tarakanov et al, Springer Verlag, 2003.
- Proceedings of the International Conference on Artificial Immune Systems (ICARIS), Springer Verlag, 2003.

- Artificial Immune Systems Forum Webpage: http://www.artificial-immune-systems.org/artist.htm
- Artificial Immune Systems Bibliography: http://issrl.cs.memphis.edu/ Artificial Immune Systems/Artificial Immune Systems_bibliography.pdf